%% file: main.tex
\documentclass[letterpaper, 10 pt, conference]{ieeeconf}  

\IEEEoverridecommandlockouts                              

\title{\LARGE \bf
Dynamic Modeling of Hand-Object Interactions via Tactile Sensing
}

\author{
Qiang Zhang$^*$,
Yunzhu Li$^*$,
Yiyue Luo,
Wan Shou,
Michael Foshey,\\
Junchi Yan,
Joshua B. Tenenbaum,
Wojciech Matusik,
and Antonio Torralba
\thanks{*Equal contribution.
Q. Zhang and J. Yan are with Shanghai Jiao Tong University. Y. Li, Y. Luo, W. Shou, M. Foshey, J. B. Tenenbaum, W. Matusik, and A. Torralba are with the Computer Science and Artificial Intelligence Laboratory (CSAIL) at Massachusetts Institute of Technology.
}
}

\usepackage{etoolbox}
\makeatletter
\patchcmd{\@makecaption}
  {\scshape}
  {}
  {}
  {}
\makeatother

\usepackage{times}
\usepackage{epsfig}
\usepackage{graphicx}
\usepackage{amsmath}
\usepackage{amssymb}

\usepackage{tabulary,multirow,multicol,overpic}
\usepackage{booktabs}

\usepackage{adjustbox}
\usepackage{array}
\usepackage{booktabs}
\usepackage{colortbl}
\usepackage{float,wrapfig}
\usepackage{hhline}
\usepackage{multirow}
\usepackage{subcaption} 
\usepackage[font={small}]{caption}

\usepackage{cuted}
\usepackage{capt-of}
\usepackage{color}
\usepackage{xcolor}

\input{math_commands.tex}

\begin{document}

\maketitle
\thispagestyle{empty}
\pagestyle{empty}

\input{text/abstract.tex}
\input{text/intro.tex}

\input{text/related.tex}

\input{text/method.tex}

\input{text/experiment.tex}

\input{text/conclusion.tex}
\bibliographystyle{IEEEtran}
\bibliography{IEEEexample}

\end{document}

%% file: math_commands.tex
\usepackage{amsmath,amsfonts,bm}

\def\eqref#1{equation~\ref{#1}}

\def\1{\bm{1}}

\def\va{{\bm{a}}}
\def\vb{{\bm{b}}}

\def\ve{{\bm{e}}}

\def\vg{{\bm{g}}}

\def\vq{{\bm{q}}}

\def\vs{{\bm{s}}}

\DeclareMathAlphabet{\mathsfit}{\encodingdefault}{\sfdefault}{m}{sl}
\SetMathAlphabet{\mathsfit}{bold}{\encodingdefault}{\sfdefault}{bx}{n}



%% file: text/abstract.tex
\input{figText/teaser}

\begin{abstract}

Tactile sensing is critical for humans to perform everyday tasks. While significant progress has been made in analyzing object grasping from vision, it remains unclear how we can utilize tactile sensing to reason about and model the dynamics of hand-object interactions. In this work, we employ a high-resolution tactile glove to perform four different interactive activities on a diversified set of objects. We build our model on a cross-modal learning framework and generate the labels using a visual processing pipeline to supervise the tactile model, which can then be used on its own during the test time. The tactile model aims to predict the 3d locations of both the hand and the object purely from the touch data by combining a predictive model and a contrastive learning module. This framework can reason about the interaction patterns from the tactile data, hallucinate the changes in the environment, estimate the uncertainty of the prediction, and generalize to unseen objects. We also provide detailed ablation studies regarding different system designs as well as visualizations of the predicted trajectories. This work takes a step on dynamics modeling in hand-object interactions from dense tactile sensing, which opens the door for future applications in activity learning, human-computer interactions, and imitation learning for robotics.

\end{abstract}

%% file: figText/teaser.tex
\begin{strip}
\centering
\vspace{-60pt}
\includegraphics[width=1.0\linewidth]{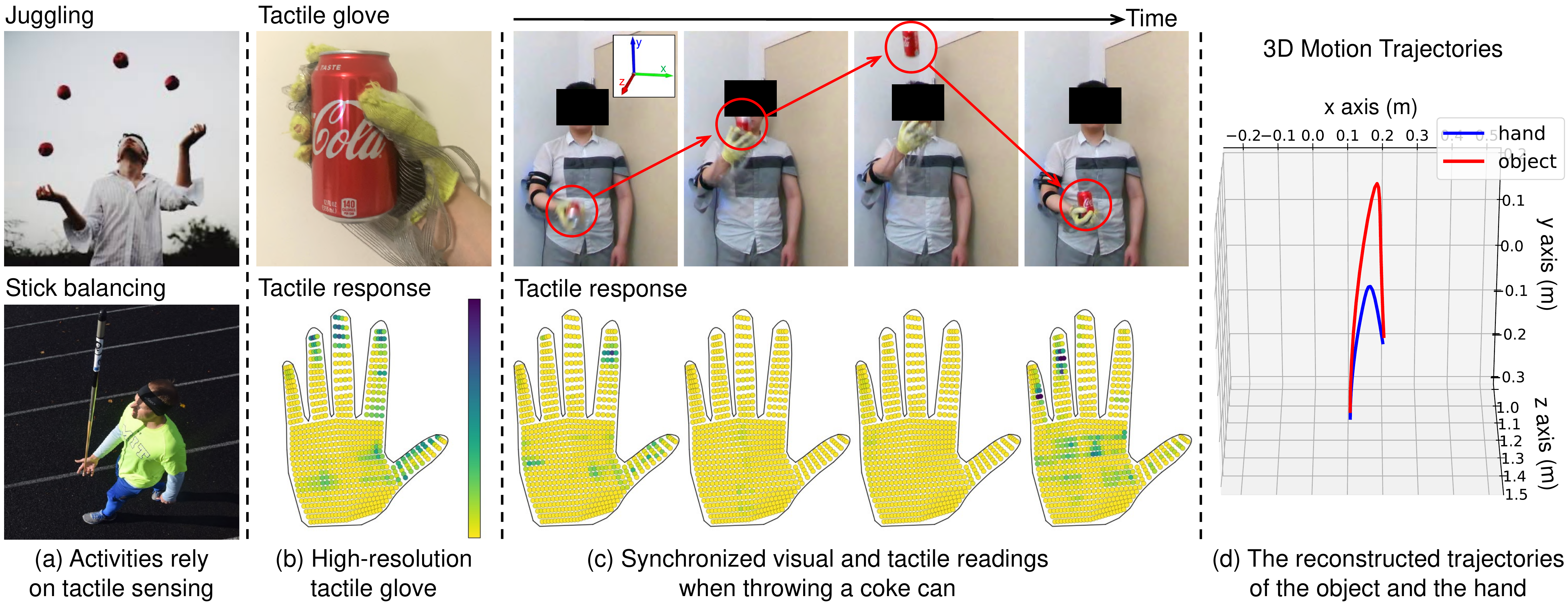}
\vspace{-15pt}
\captionof{figure}{\textbf{Hand-Object Interactions Using a Tactile Glove.} (a) Daily activities involving hand-object interactions require tactile sensing and an intuitive understanding of real-world dynamics. (b) We employ a high-resolution tactile glove~\cite{sundaram2019learning} to obtain the tactile signal when interacting with objects. (c) When throwing an object in the air and catching it, different tactile patterns appear during the process. (d) We model the dynamics of this interaction and reconstruct the 3D trajectories of both the object and the hand using only tactile data.}
\label{fig:teaser}
\vspace{-5pt}
\end{strip}

%% file: text/intro.tex
\section{Introduction}

When interacting with diverse objects using our hands, humans can feel and sense the contacts, enabling a strong intuitive understanding of the dynamics of the interactions (Figure~\ref{fig:teaser}a).
This understanding allows us to predict how the environment will evolve and be influenced by these interactions, which is tremendously helpful for accomplishing daily manipulation tasks.
To further our understanding of human activities and make progress in robotics, it is desirable to develop techniques that can model the dynamics of hand-object interactions, especially in highly dynamic scenarios.
In recent years, there has been a growing amount of work aiming to understand and reconstruct hand-object interactions. Some of these works estimate the pose or reconstruct the shapes of both the objects and the hand~\cite{glauser2019interactive,hasson2019learning}
, while others take a step further by modeling the contact when grasping an object~\cite{bernardin2005sensor,brahmbhatt2019contactdb,brahmbhatt2020contactpose}. 
However, most of the works focus on static grasps due to the reliance on indirect measurements such as thermal cameras to estimate the contact regions. The inability to provide an on-the-fly estimation of the contacts and the applied forces limits the use of these methods in more dynamic scenarios.

To directly sense the contacts, many researchers have built tactile sensors and explored how tactile sensing can recognize objects~\cite{gandarias2019cnn,corradi2015bayesian,bauza2019tactile}, classify materials~\cite{erickson2017semi,zheng2019bio,wang2020recognition}, and how it can be coupled with the vision for representation learning~\cite{Yuan_2017_CVPR,Li_2019_CVPR,lee2019touching}.
Some prior works have tried to model the dynamics of the tactile interaction~\cite{tian2019manipulation,she2019cable,narang2020interpreting}, yet most of them only use tactile sensing at the fingertip area and primarily focus on small motion.
In this work, we aim to model the dynamics of hand-object interactions in highly dynamic tasks using a high-resolution tactile glove developed by Sundaram~et~al.~\cite{sundaram2019learning}~(Figure~\ref{fig:teaser}b).
Specifically, we employ a cross-modal learning framework that we collect synchronized RGBD images and tactile readings. During training, we use a visual processing pipeline to estimate the ground truth trajectories of both the object and the hand, which act as the labels to supervised the tactile model. During the test time, we only assume to have access to the tactile inputs and predict the corresponding system's states using the trained tactile model.
The most desirable scenario would be to have a dexterous robotic hand performing the interactive tasks.
However, although less accurate at repeating a specific action sequence, human capabilities are still far ahead of the current state-of-the-art robotic systems~\cite{kim2014catching,akkaya2019solving,andrychowicz2020learning,nagabandi2020deep}.
Therefore, we ask a human subject to wear the glove to perform interactive activities in order {\it not} to limit ourselves from studying more dexterous and dynamic interactions with the objects.
We consider four activities:
(1) holding an object and waving it in the air,
(2) throwing an object upwards and catch it~(Figure~\ref{fig:teaser}c),
(3) bouncing a ping-pong ball up and down using the racket, and
(4) balancing an umbrella to prevent it from falling. 
Our goal is to predict the motion trajectory of the hand and the object purely based on the tactile readings~(Figure~\ref{fig:teaser}d).

Our tactile model builds on top of a predictive model that takes the current state of the system, i.e., the positions and velocities of the object and the hand, and the tactile reading as input, then predicts the system state at the next time step. We add to the model an object encoder that produces object-centric embeddings calculated from a short sequence of interaction data that implicitly encodes the object-specific information like the mass and geometry.
Experiments show that our model can reason about the interaction patterns from the tactile data, model the dynamics of the interactions, and quantify the uncertainty of the prediction. The use of the object encoder also allows an improved performance when generalizing to unseen objects. We also provide detailed ablation studies regarding the resolution of the tactile sensing and the visualizations of the predicted trajectories.
This work moves in the direction of dynamics modeling in hand-object interactions from dense tactile sensing, which we hope can facilitate future studies and applications in multiple domains, including computer vision, graphics, and robotics.

%% file: text/related.tex
\input{figText/model.tex}

\section{Related Work}

\subsubsection{Tactile sensing for perception and control}
There has been a long history of hardware design for tactile sensing~\cite{cutkosky2008force,lederman2009haptic,johnson2009retrographic,johnson2011microgeometry}.
Kappassov~et~al.~\cite{kappassov2015tactile} provide an excellent summary of tactile sensors for dexterous robot hands. Recently, Yuan~et~al.~\cite{yuan2017gelsight} propose to use a vision-based optical tactile sensor, which enables high-fidelity measurements of the contacts and object properties~\cite{yuan2016estimating,yuan2017shape}.
Liu~et~al.~\cite{liu2017glove,liu2019high} build a tactile glove for studying object grasping and hand-object manipulation.
Luo~et~al.~\cite{luo2021learning} propose to build garments consist of conformal tactile textiles to learn human-environment interactions.
In this paper, we use the tactile glove with a much higher resolution developed by Sundaram~et~al.~\cite{sundaram2019learning}.
The glove is flexible and deformable, giving us dense tactile sensing covering the palm area, allowing us to record tactile information while performing various interactive tasks with diverse objects.
With the help of these tactile sensors, previous researches improve the performance of various control tasks, ranging from basics object manipulation~\cite{lee2019making,tian2019manipulation}, slip control~\cite{veiga2018grip,dong2019maintaining}, to regrasping~\cite{chebotar2016self,hogan2018tactile} and others.
This paper shares a similar spirit with~\cite{tian2019manipulation} that learns to model the dynamics with tactile sensing, but mainly focus on hand-object interactions using a high-resolution tactile glove.

\subsubsection{Modeling across multiple modalities}
Frome~et~al.~\cite{frome2013devise} seek to find the correlation between vision and language.
Other researches also explored image captioning~\cite{karpathy2015deep,xu2015show} and learning image-text joint embeddings~\cite{norouzi2013zero,otani2016learning,aytar2017cross}. 
Finding the relationship between vision and sound is another interesting topic and has drawn increasing attention from the community~\cite{owens2016ambient,gan2020foley,zhao2019sound}.
Researchers have also proposed to connect vision and touch via cross-domain modeling~\cite{Yuan_2017_CVPR,falco2017cross,Li_2019_CVPR,lee2019touching} or estimating 3D human poses from a tactile carpet by taking the outputs of computer vision models as supervisions~\cite{luo2021intelligent}.
Others have tried to combine tactile sensing with proprioception and kinesthetic information for object and shape recognition~\cite{luo2019iclap,pastor2020bayesian}.
In this paper, we also desire to establish a connection between touch and vision but focusing on the task of dynamics modeling using the newly-developed tactile glove~\cite{sundaram2019learning}.

\subsubsection{Hand-object interactions}
Recent advances in learning human-object interactions focus on estimating and tracking the hand in hand-object interaction scenarios~\cite{hasson2019learning,oikonomidis2011full,sridhar2016real},
where most of the works take images as the primary sensing inputs~\cite{kyriazis2014scalable,pham2015towards,tzionas2016capturing}.
Estimating and reasoning about the contacts is critical for building a better model of the interactions~\cite{bernardin2005sensor,sundaram2019learning}.
Therefore, Brahmbhatt~et~al.~\cite{brahmbhatt2019contactdb,brahmbhatt2020contactpose} use a thermal camera to label the contact regions during grasping.
Some recent works~\cite{pham2015towards,pham2017hand,ehsani2020use} aim to infer contact force from visual images or take a step further by predicting the motion trajectories.
Overall, prior methods rely on indirect measurements or weak supervision from a physics-based simulator and do not directly use real tactile data.

\subsubsection{Contrastive learning}
People have been using triplet loss for better representation learning~\cite{chopra2005learning,schroff2015facenet},
while others have investigated the use of contrastive loss~\cite{hadsell2006dimensionality}.
Recently, many papers use contrastive learning to get better self-supervised pre-trained features to boost the performance of downstream tasks~\cite{gutmann2010noise,wu2018unsupervised,oord2018representation}.
Other researchers have been using contrastive learning to find better representations via multi-task learning, e.g., Sun~et~al.~\cite{sun2014deep} combine softmax and contrastive loss to help face recognition tasks.
Zhang~et~al.~\cite{zhang2016embedding} use triplet loss to get better features by embedding label structures.
In this project, we use the contrastive loss to facilitate the model's generalization performance in multi-object scenarios, where we hope to learn an embedding space that contains object-specific information like the mass and the geometry.

%% file: figText/model.tex
\begin{figure*}[!t]
\centering
\includegraphics[width=.85\linewidth]{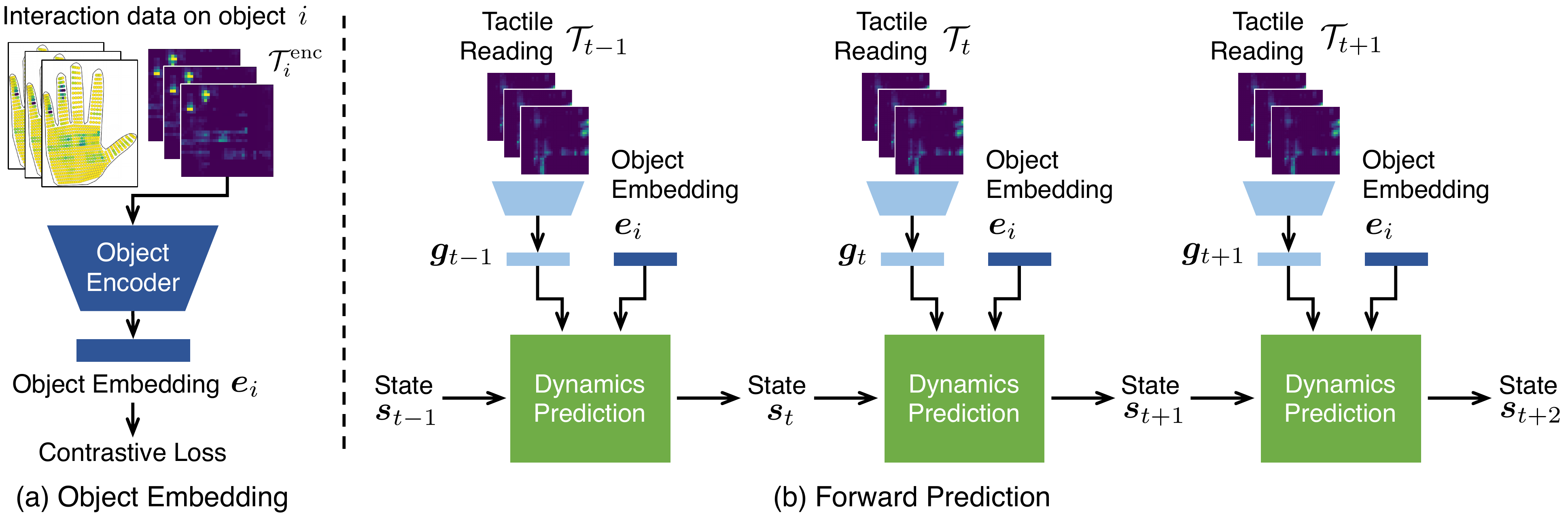}
\vspace{-3pt}
\caption{\textbf{Dynamics Prediction Model.}
(a) The object embedding $\ve_i$ is derived by encoding a short period of interaction data into an implicit object-specific representation containing information like the mass and geometry. (b) Our model builds on top of a dynamics prediction model, which takes (i) the current state $\vs_t$, i.e., the positions and velocities of the object and the hand, (ii) the tactile embedding $\vg_t$, and (iii) the object embedding $\ve_i$ as input, and then predicts the system state at the next step.
}
\label{fig:method}
\vspace{-10pt}
\end{figure*}

%% file: text/method.tex
\section{Method}
\label{sec:method}

In this section, we present the details of our dynamics modeling framework. We first discuss some details of the tactile glove and go through the overall structure of our dynamics module in Section~\ref{sec:dynamics}. We then describe how we quantify the uncertainty using pinball loss (Section~\ref{sec:pinball}) and improve the generalization performance by extracting an object embedding using contrastive learning (Section~\ref{sec:contrast}).
In Section~\ref{sec:metrics}, we describe our overall loss and how we measure the performance of the dynamics modeling.

\subsection{Tactile Glove.}
\label{sec:glove}

We use the tactile glove proposed by Sundaram~et al.~\cite{sundaram2019learning} to collect the hand-object interaction touch data (Figure~\ref{fig:teaser}b). There are about $680$ sensors covering the palm area of the glove that can measure the applied forces.
We record the tactile signal at each time step as an image of $32\times32$ resolution at a rate of around $15$ Hz.

\subsection{Dynamics Prediction Model.}
\label{sec:dynamics}

The overall dynamics model is defined as the following:
\begin{equation}
    \hat{\vs}_{t+1} = f(\vs_t, \vg_t, \ve_i),
\end{equation}
where $\vs_t$ denotes the state of the system at time $t$, $\vg_t$ indicates the tactile embedding derived from the current tactile input, and $\ve_i$ is the object embedding that is shared across all time steps, containing contextual information like the mass and geometry of object $i$.

We represent the state of our system using the states of both the object and the hand, $\vs_t = (\vs^\text{object}_t, \vs^\text{hand}_t)$, each of which is a concatenation of its position and velocity,
\begin{equation}
\begin{aligned}
    \vs^\text{hand}_t = (\vq^\text{hand}_t, \dot{\vq}^\text{hand}_t), \quad
    \vs^\text{object}_t = (\vq^\text{object}_t, \dot{\vq}^\text{object}_t),
\end{aligned}
\end{equation}
where $\vq_t = (x_t, y_t, z_t)$ indicates its 3D location.

To calculate the tactile embedding $\vg_t$, we take a small temporal window of $M$ tactile frames as input, $\mathcal{T}_t = \{T_{t-M+1}, \dots, T_{t-1}, T_t\}$, concatenate them to form a tensor of shape $M\times32\times32$, and then feed it through an encoder (Figure~\ref{fig:method}b). We implement the encoder as a three-layer convolutional network with channel sizes of 32, 64, 128, individually.
The dynamics prediction module (the green box in Figure 2b) is an MLP (Multi-Layer-Perceptron) that predicts the next state $\hat{\vs}_{t+1}$ by taking as input the concatenation of the tactile embedding $\vg_t$, the object embedding $\ve_i$, and the system state $\vs_t$ at the last time step.

\subsection{Pinball Loss for Uncertainty Estimation.}
\label{sec:pinball}

To quantify the uncertainty in state prediction, we use a pinball loss function~\cite{koenker2001quantile} to predict both the mean value and the $10\%$ and $90\%$ quantiles. The intuition is that for a given set of inputs, the model estimates through a single forward pass both the expected future state and a cone of error within which the ground truth should lie $80\%$ of the whole time.

More specifically, the outputs of the dynamics prediction module contain both the state $\hat{\vs}$ and the offset from the expected state, $\bm{\sigma}$, such that $\hat{\vs}+\bm{\sigma}$ is the $90\%$ quantile of its distribution while $\hat{\vs}-\bm{\sigma}$ is the $10\%$ quantile. Assuming that the ground truth future state is $\vs$, the pinball loss is calculated as the following:
\begin{equation}
\begin{aligned}
\mathcal{L}_\text{pinball}^{\tau} & = \max(\tau \hat{p}_{\tau},-(1-\tau)\hat{p}_{\tau}), \text{where} \\
\hat{p}_{\tau} & =
  \begin{cases}
        \vs-(\hat{\vs}-\bm{\sigma)}, \quad \text{for } \tau \leq 0.5\\
        \vs-(\hat{\vs}+\bm{\sigma}), \quad \text{otherwise.}
  \end{cases}
\end{aligned}
\end{equation}
We use the L1 regression loss and this pinball loss for both the predicted state and the quantiles at $\tau=0.1$ and $\tau=0.9$, making $\bm{\sigma}$ a measurement of the difference between the $10\%$
and $90\%$ quantiles and the expected value. The final loss for future state prediction is then:
\begin{equation}
    \mathcal{L}_\text{state} = | \vs - \hat{\vs} | + \mathcal{L}_\text{pinball}^{0.1} + \mathcal{L}_\text{pinball}^{0.9}.
\end{equation}

\input{figText/dataset.tex}

\subsection{Contrastive Learning for Object Embeddings.}
\label{sec:contrast}

To improve our model's generalization ability to unseen objects, we propose to encode a short period of interaction data into an implicit object embedding containing object-specific information like the mass and geometry.
The intuition here is that when we humans have interacted with an object for a while using our hands, we can develop a better intuitive understanding of the object's dynamics and give better estimates of the object's physical properties like mass.

We subsample a short sequence from the interaction dataset of the object $i$ and denote it as $\mathcal{T}^\text{enc}_i$. An encoder, implemented as a convolutional neural network, maps $\mathcal{T}^\text{enc}_i$ into its object embedding $\ve_i$ (Figure~\ref{fig:method}a).
When predicting multiple steps into the future using the dynamics prediction module, the tactile embedding $\vg_t$ is changing as it is derived from the current tactile inputs.
In contrast, the object embedding $\ve_i$ is kept the same throughout the prediction horizon~(Figure~\ref{fig:method}b), which we hope can help the model disentangle the object-related information from the motion-related features, facilitating generalization to unseen objects.

Inspired by recent developments in contrastive representation learning~\cite{oord2018representation}, we introduce an auxiliary contrastive loss to help the object encoder learn more meaningful representations. We use the object embeddings calculated from the same object as the positive pair and embeddings from different objects as the negative pair. We then define the auxiliary contrastive loss as the following:
\begin{equation}
\mathcal{L}_\text{obj}=\mathbb{E}_\ve \left[\log \frac{\exp(\ve^T\cdot \ve^+)}{\exp(\ve^T\cdot \ve^+)+\Sigma_{j=1}^{N_\text{sample}-1}\exp(\ve^T\cdot \ve^-)}\right],
\end{equation}
where $\ve$ is the reference embedding, $\ve^+$ and $\ve^-$ are positive and negative embeddings respectively, and $N_\text{sample}$ denotes the number of samples.

\subsection{Overall Loss and Evaluation Metrics.}
\label{sec:metrics}

The overall loss of our framework is then a weighted sum of the future state prediction loss $\mathcal{L}_\text{state}$ and the contrastive loss for learning the object embeddings $\mathcal{L}_\text{obj}$:
\begin{equation}
    \mathcal{L} = \mathcal{L}_\text{state} + \lambda_\text{obj}\mathcal{L}_\text{obj}.
    \label{eqn:loss}
\end{equation}

We evaluate the performance of our dynamics modeling framework using two different metrics: (1) the L1 distance between the predicted state and the ground truth over different time steps, and (2) the Hausdorff distance~\cite{huttenlocher1993comparing} between the predicted and ground truth trajectories.
Given the ground truth trajectory $A = \{ \vs_t, \vs_{t + 1}, ... \}$ and the predicted trajectory $B = \{ \hat{\vs}_t, \hat{\vs}_{t + 1}, ... \}$, the Hausdorff distance measures the similarity based on the shape of the trajectories and is defined as the following:
\begin{equation}
    H=\max \{\max_{\va_i\in A} \min_{\vb_j\in B}d(\va_i,\vb_j),\max_{\vb_i\in B} \min_{\va_j\in A}d(\vb_i,\va_j)\},
\end{equation}
where $d(\cdot, \cdot)$ denotes the Euclidean distance.

%% file: figText/dataset.tex
\begin{figure*}[!t]
\centering
\includegraphics[width=1.0\linewidth]{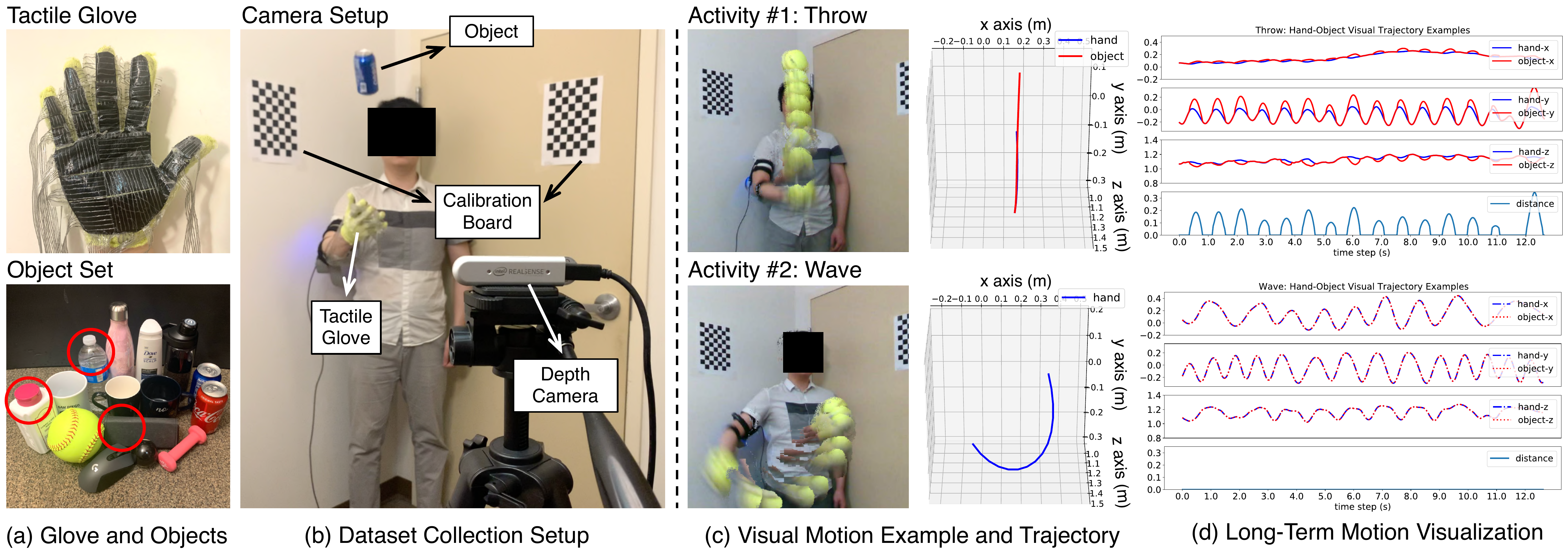}
\vspace{-17pt}
\caption{\textbf{Data Collection Setup and Motion Examples.}
(a) The tactile glove and the object set for performing ``wave" and ``throw" activities.
Objects used for evaluating the model's generalization ability are marked using red circles (unseen during training).
(b) A depth camera collects RGBD videos, which we use to project the hand and object's motion to 3D. (c) Here we show a visual example and the corresponding 3D trajectory for each activity. (d) shows long-term visualizations of typical motions in each axis and the distance between hand and object for the two activities.
}
\label{fig:dataset}
\vspace{-10pt}
\end{figure*}

%% file: text/experiment.tex
\section{Experiments}

\input{tabText/dataset.tex}

\subsection{Data Collection Setup.}

\input{figText/quali}

The primary focus of this work is the dynamics modeling of hand-object interactions from tactile sensing.
Given that there is still a huge performance gap between the state-of-the-art dexterous hand robotic system~\cite{kim2014catching,akkaya2019solving,andrychowicz2020learning,nagabandi2020deep} and human capabilities, we ask a human subject to wear the glove and perform various interaction tasks instead of using a robot system, which allows us to study more dexterous and dynamic interactions.

Figure~\ref{fig:dataset}b shows the setup for data collection, where a human subject faces a depth camera and performs various hand-object interaction activities.
The RGBD camera and the tactile glove collect synchronized visual and tactile data with a framerate of 20 Hz and 15 Hz, respectively.
To obtain the ground truth locations of both the hand and the object, we first segment the image according to the depth map and use the CSRT tracker~\cite{Lukezic_IJCV2018} implemented in the OpenCV library to track their locations.
We also calibrate the camera to ensure that we can interpret the ground truth labels in real physical units like meter and second.
Compared with motion capture systems using markers on the object/hand, this RGBD-based caption system is simple, cheap, and easy to use.
See the accompanying videos for visualizations of the tracking results.

We first consider the {\it Wave} activity, where the human subject holds an object in hand and wave it in the air.
This activity is to verify the feasibility of the task: only when there exists a strong correlation between the tactile signal and the object's acceleration can we predict the motion trajectories purely from touch.
This intuition on the correlation between different sensory modalities originates from Newton's second law of motion, which builds the connection between force (perceived by the tactile signal) and motion (the object's acceleration).
Then we extend the system to more complicated activities where there can be large relative motion between the object and the hand:
\begin{itemize}
\item {\it Throw}: throw an object upward and catch it,
\item {\it PingPong}: bounce a ping-pong ball using the racket,
\item {\it Balance}: balance an umbrella to prevent it from falling.
\end{itemize}
Examples of the four activities are shown in Figure~\ref{fig:dataset} and Figure~\ref{fig:newtask} respectively.
For {\it Wave} and {\it Throw}, we collect data for interacting with 15 different objects~(Figure~\ref{fig:dataset}a), where we constantly change the in-hand position of the objects to increase the diversity of the dataset.
We then split them into 12 seen and 3 unseen objects.
Among the data for the seen objects, $80\%$ of them are used for training, while the remaining $20\%$ are for evaluation. The detailed statistics are shown in Table~\ref{tab:stat0}.

\subsection{Experimental Results on ``Wave" and ``Throw".}

This section presents the experimental results on the {\it Wave} and {\it Throw} tasks. We first evaluate the model's performance by training separate models for each object.
Then, we train a single model on all objects to test the model's multi-object generalization ability and show the benefits brought by the object embedding.

\subsubsection{Results on Single Object.}

We train a separate model for each object by minimizing the loss fucntion defined in Equation~\ref{eqn:loss} for $50$ epochs using Adam optimizer~\cite{kingma2014adam} and a learning rate of $0.0005$. 

We compare our model with the following two baselines:
(1) Random: For each testing example, we randomly sample another trajectory from the dataset and measure their difference using the metrics discussed in Section~\ref{sec:metrics}, which acts as a lower bound for the performance.
(2) w/o Touch: In this baseline, we remove the tactile input and only use the state at the previous time step ($\vs_t$ in Figure~\ref{fig:method}b) as the input to the dynamics prediction module. This baseline serves to show the benefit of the tactile information in dynamics modeling.

The left part of Table~\ref{tab:exp_merge} illustrates the quantitative results. It is clear from the table that the touch information is critical for predicting the dynamics.
We also include some qualitative results in Figure~\ref{fig:quali} for each activity by showing a side-by-side illustration of the original RGB image, the motion trajectories in 3D, and the estimated uncertainty in each axis.
Our model can generate trajectories that closely match the ground truth, where the hand and object's movements consistently lie in our model's predicted confidence interval.

\input{tabText/quant.tex}
\input{figText/newtask.tex}
\input{figText/ablate_res.tex}

\subsubsection{Results on Multi-Object Generalization}
This section evaluates the model's ability to generalize across multiple objects. We use an object encoder to extract the implicit embeddings for each object that contain object-related information like the mass and the geometry (Figure~\ref{fig:method}a).
To optimize our model, we first train the object encoder using contrastive loss introduced in Section~\ref{sec:contrast}, where the positive and negative samples are from the same object and different objects separately.
Each sample is a $20$-step touch segment uniformly sampled from all touch data sequences.
We then fix the object embedding for each object and concatenate it with the tactile embedding as the input feature for the subsequent computation (Figure~\ref{fig:method}b).
During the test time, we choose the touch sequence from the first $20$ steps and feed it into the object encoder to obtain the object embeddings.

We compare our model with a baseline that does not use the object embedding (w/o Embedding). 
The right half of Table~\ref{tab:exp_merge} presents the quantitative results on cross-object generalization performance, which shows the benefit of the object embedding for improving the generalization ability.

\subsubsection{Ablation Study on the Tactile Resolution}
In this section, we aim to evaluate the importance of sensor resolution in dynamics modeling.
We downsample the input tactile image from original $32\times32$ resolution to $24\times24$, $16\times16$, $12\times12$, $8\times8$, and $6\times6$ respectively, and evaluate the performance on the ``Billiard" dataset from the ``Waving" task.
As shown in Figure~\ref{fig:ablation}, with the decrease of the effective resolution, the prediction performance drops notably, showing the importance of the sensor count in dynamics modeling.

\subsection{Results on ``PingPong" and ``Balance".}

We also evaluate our model's performance on the two other hand-object interaction activities: {\it PingPong} and {\it Balance}.
The training and test numbers are 15k and 3k, respectively, for both of these two activities.
Our model takes the tactile sequences as input and predicts the movements of the ping-pong ball and the umbrella.
In {\it PingPong}, the state of the object $\vs^\text{object}_t$ is represented as the position and velocity of the ping-pong ball, while in {\it Balance}, we use the position and velocity of the umbrella's upper tip as the object state.
The quantitative results and the qualitative results are shown in Table~\ref{tab:others} and Figure~\ref{fig:newtask}, respectively.

\input{tabText/newtask.tex}

%% file: tabText/dataset.tex
\begin{table}[!t]

\begin{center}
\begin{tabular}{cccccc}
\toprule
\multicolumn{1}{l}{Dataset}  & Split & Object \# & Train  & Eval  & All    \\ \hline
\multirow{2}{*}{Wave}  & Seen    & 12       & 166k & 42k & 208k \\
                       & Unseen  & 3        & 0      & 58k & 58k  \\ \hline
\multirow{2}{*}{Throw} & Seen    & 12       & 184k & 46k & 230k \\
                       & Unseen  & 3        & 0      & 61k & 61k \\ \hline
\end{tabular}
\end{center}
\vspace{-8pt}
\caption{\textbf{Dataset Statistics.}
There are 12 seen objects and another 3 unseen objects (marked with red circle in Figure~\ref{fig:dataset}a) to evaluate the model's generalization ability. For the seen object dataset, 80\% is kept for training while the remaining is for evaluation. For the unseen object, all of the data is reserved for evaluation.
}
\label{tab:stat0}
\vspace{-10pt}
\end{table}

%% file: figText/quali.tex
\begin{figure*}[!t]
\centering
\includegraphics[width=1.0\linewidth]{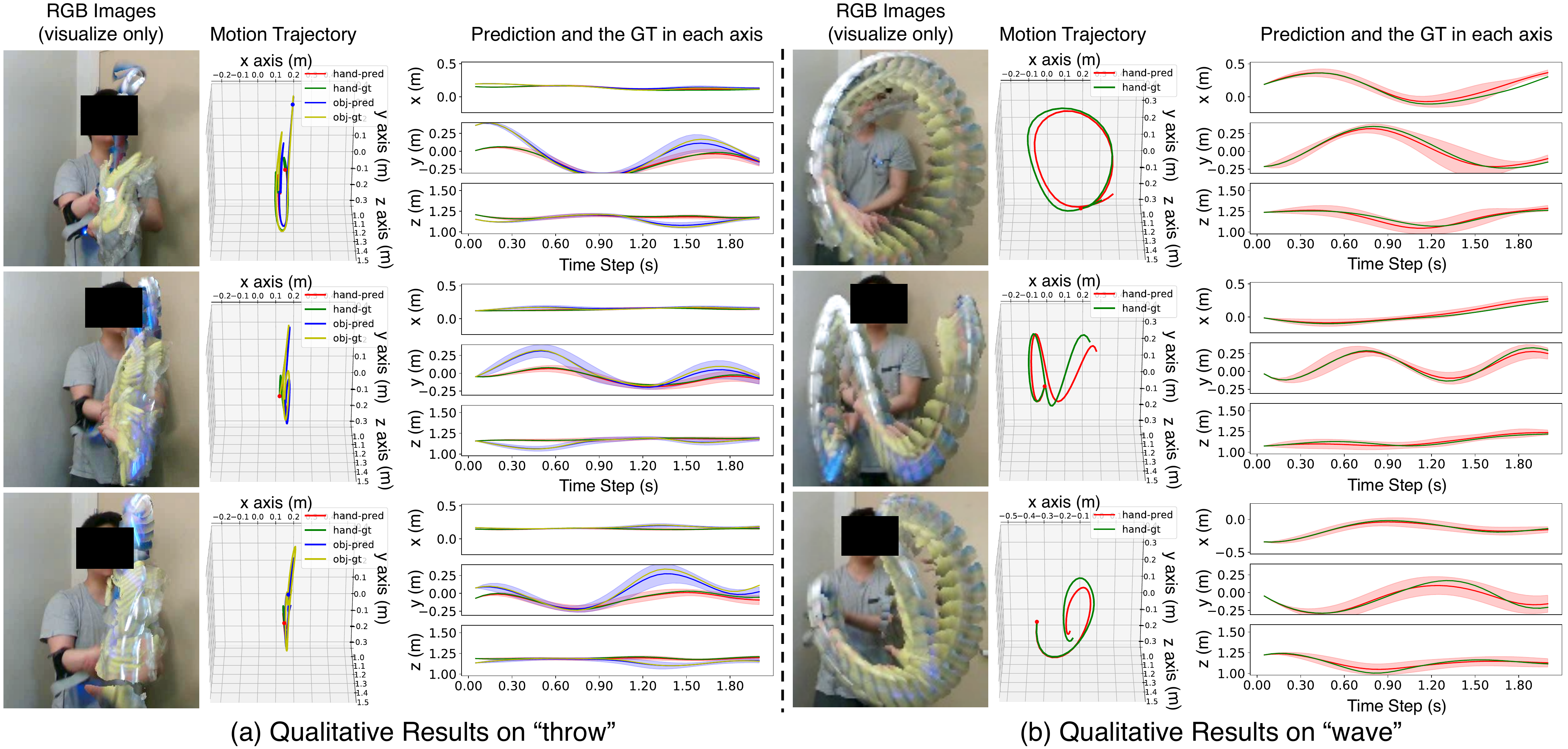}
\vspace{-15pt}
\caption{\textbf{Qualitative Visualizations for ``Throw" and ``Wave".}
(a) Visualization of the results in the ``throw" activity. From left to right, we show the reference images (only used for visualization), the predicted and the ground truth trajectory in 3D, and the individual comparison in each axis. 
The shaded area indicates the estimated uncertainty from our model that corresponds to the $10\%$ and $90\%$ quantiles.
(b) The same set of visualizations for the ``wave" activity. Our model's predictions closely resemble what's happened in the real physical world.
}
\label{fig:quali}
\vspace{-10pt}
\end{figure*}

%% file: tabText/quant.tex
\begin{table*}[h]
\begin{center}
\begin{tabular}{ccccccccc}
\toprule
Exp. type  & \multicolumn{3}{c|}{Single object}               & \multicolumn{4}{c}{Multi-object}                                                  \\ \hline
Setting   & \multicolumn{3}{c|}{Average}                     & \multicolumn{2}{c|}{Seen}                          & \multicolumn{2}{c}{Unseen}    \\ \hline
Throw     & Random & w/o Touch & \multicolumn{1}{l|}{Ours} & w/o Embedding & \multicolumn{1}{l|}{Ours} & w/o Embedding & Ours \\ \hline
L1@20    &   0.992 &0.341 & \textbf{0.227}   &  0.281  & \textbf{0.254}   & 0.312 &  \textbf{0.265}  \\
L1@50    &  1.005 & 0.547 & \textbf{0.288} & 0.323   & \textbf{0.297}   &  0.391   &\textbf{0.305}\\
Hausdorff & 1.169  & 1.007 &\textbf{0.659} & 0.868  & \textbf{0.775}  & 0.918 & \textbf{0.842}  \\
\hline
Wave      & Random & w/o Touch  & \multicolumn{1}{l|}{Ours} & w/o  Embedding & \multicolumn{1}{l|}{Ours} & w/o Embedding & Ours \\ \hline
L1@20    & 0.308 & 0.186  & \textbf{0.108}  & 0.135  & \textbf{0.116}    &  0.154            & \textbf{0.129}     \\
L1@50    & 0.309 &  0.212 & \textbf{0.132} &  0.161  & \textbf{0.137}   &  0.182   &  \textbf{0.154}    \\
Hausdorff & 1.671 & 1.498 & \textbf{0.917}  &  1.154   & \textbf{1.025}   &  1.308     & \textbf{1.119} \\
\hline
\end{tabular}
\end{center}
\vspace{-8pt}
\caption{\textbf{Quantitative Results for ``throw" and ``wave".}
Here we assess our model's ability for predicting the dynamics of a single object and that of cross-object generalization.
We evaluate the performance using the Hausdorff distance as well as the average L1 distance between the prediction and the ground truth when rolling 20 or 50 steps into the future.
In both the ``throw" and the ``wave" activities, our model outperforms a random baseline and a baseline that does not use the touch information, which is in line with our intuition that touch plays a critical role in modeling the dynamics.
The use of object embeddings also improves the performance, especially when encountering objects that are not seen during training.
}
\label{tab:exp_merge}
\vspace{-10pt}
\end{table*}

%% file: figText/newtask.tex
\begin{figure*}[htbp]
\centering
\includegraphics[width=\linewidth]{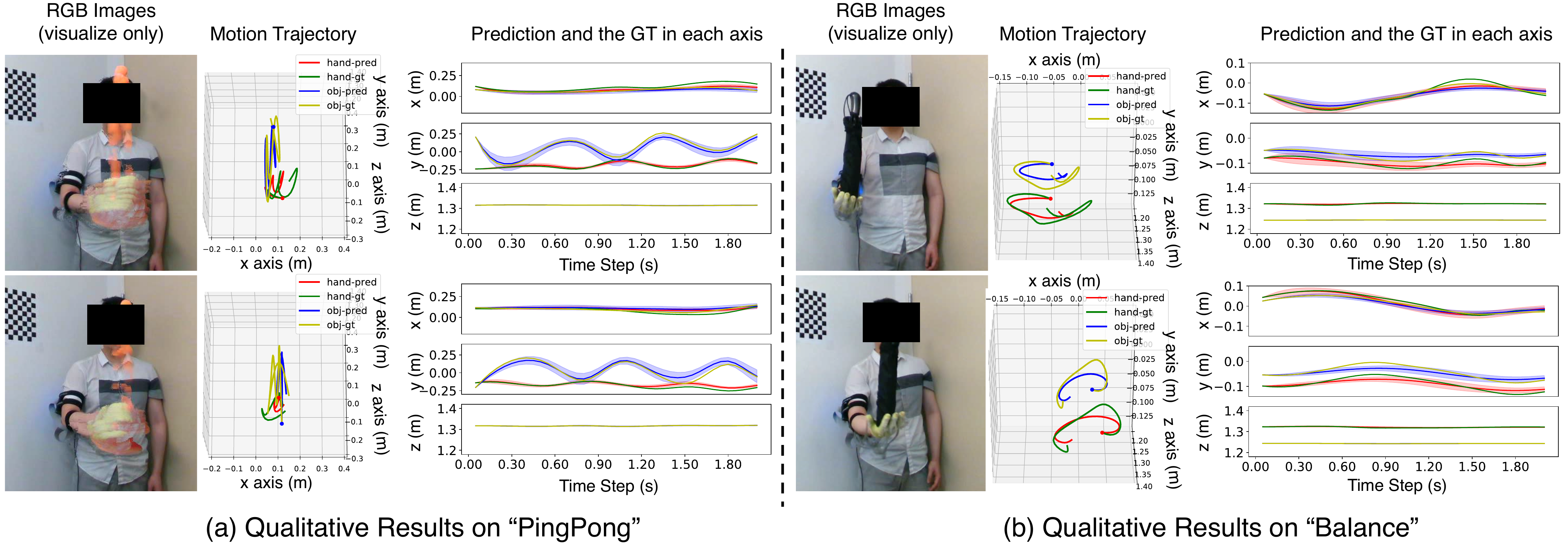}
\vspace{-15pt}
\caption{\textbf{Qualitative Visualizations for ``Ping-Pong" and ``Balance".}
Similar to Figure~\ref{fig:quali}, we show the reference images, the predicted and the ground truth 3D trajectory, and individual comparisons in each axis.
The state of the object $\vs^\text{object}_t$ is represented as the position and velocity of the ping-pong ball and the upper tip of the umbrella in these two activities, respectively.
Again, our model's prediction closely matches the evolution of the actual physical systems.
}
\label{fig:newtask}
\vspace{-10pt}
\end{figure*}

%% file: figText/ablate_res.tex
\begin{figure}[h]
\centering
\includegraphics[width=.88\linewidth]{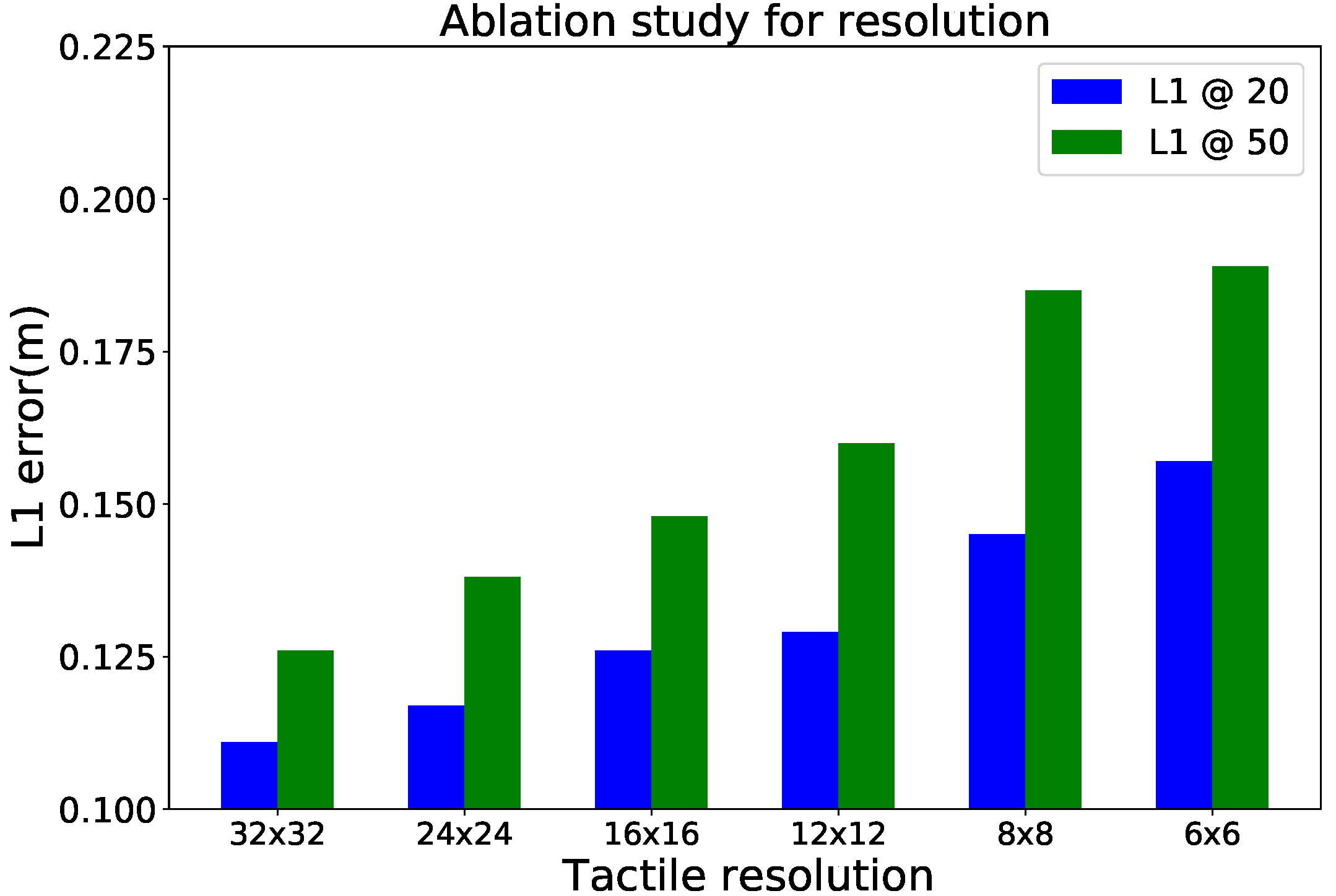}
\vspace{-3pt}
\caption{\textbf{Ablation Study on the Resolution of the Tactile Glove.}
The error bars show the L1 error of the prediction at 20 or 50 steps into the future when using tactile images of different resolutions, ranging from $32\times32$ to $6\times6$.
Higher resolution can, in general, achieve better performance.
}
\label{fig:ablation}
\vspace{-10pt}
\end{figure}

%% file: tabText/newtask.tex
\begin{table}[!t]
\begin{center}
\begin{tabular}{llll}
\toprule
PingPong  & Random & w/o Touch & Ours \\ \hline
L1@20    &  0.205   & 0.164  &  \textbf{0.092}    \\
L1@50    &  0.207   &  0.185   & \textbf{0.109}     \\
Hausdorff &  0.725   &   0.293   & \textbf{0.162} \\
\hline
Balance   & Random & w/o Touch & Ours \\ \hline
L1@20    & 0.183 &  0.139  & \textbf{0.076}  \\
L1@50    & 0.187 & 0.175 & \textbf{0.103} \\
Hausdorff & 0.835  & 0.453  & \textbf{0.267}  \\
\hline
\end{tabular}
\end{center}
\vspace{-8pt}
\caption{\textbf{Quantitative Results for ``PingPong" and ``Balance".}
Our method outperforms two baselines on all the metrics, demonstrating the effectiveness of our dynamics prediction model and the critical role tactile sensing played in the process.
}
\label{tab:others}
\vspace{-8pt}
\end{table}

%% file: text/conclusion.tex
\section{Conclusion}

In this paper, we study the problem of dynamics modeling in hand-object interactions using tactile sensing,
where a human subject acts as a proxy for a robotic system to investigate more dexterous and dynamic interactive activities.
The human subject wears a high-resolution tactile glove to perform four different interactive tasks with a range of diverse objects.
We build a model that predicts how the environment is expected to evolve with time by only taking the current state and the subsequent tactile information as input.
We further augment the model using an object-centric encoder to facilitate multi-object generalization.
Experiments show that our model can predict the dynamics of the interactions, quantify the uncertainty of the prediction, and generalize to unseen objects.
Through this work, we hope to draw people's attention and inspire future investigations to build better perception systems and intuitive models of the environment via the lens of different sensing modalities.

In the future, we desire to predict not only the position but also the orientation of the object, where we plan to combine the proposed framework with other sources of information like IMUs, or human (or robot) joint states (e.g., positions, torques), which we believe would improve and scale up the framework for more complicated applications and tasks.

%% file: main.bbl
\begin{thebibliography}{10}
\providecommand{\url}[1]{#1}
\csname url@rmstyle\endcsname
\providecommand{\newblock}{\relax}
\providecommand{\bibinfo}[2]{#2}
\providecommand\BIBentrySTDinterwordspacing{\spaceskip=0pt\relax}
\providecommand\BIBentryALTinterwordstretchfactor{4}
\providecommand\BIBentryALTinterwordspacing{\spaceskip=\fontdimen2\font plus
\BIBentryALTinterwordstretchfactor\fontdimen3\font minus
  \fontdimen4\font\relax}
\providecommand\BIBforeignlanguage[2]{{%
\expandafter\ifx\csname l@#1\endcsname\relax
\typeout{** WARNING: IEEEtran.bst: No hyphenation pattern has been}%
\typeout{** loaded for the language `#1'. Using the pattern for}%
\typeout{** the default language instead.}%
\else
\language=\csname l@#1\endcsname
\fi
#2}}

\bibitem{sundaram2019learning}
S.~Sundaram, P.~Kellnhofer, Y.~Li, J.-Y. Zhu, A.~Torralba, and W.~Matusik,
  ``Learning the signatures of the human grasp using a scalable tactile
  glove,'' \emph{Nature}, vol. 569, no. 7758, pp. 698--702, 2019.

\bibitem{glauser2019interactive}
O.~Glauser, S.~Wu, D.~Panozzo, O.~Hilliges, and O.~Sorkine-Hornung,
  ``Interactive hand pose estimation using a stretch-sensing soft glove,''
  \emph{ACM Transactions on Graphics (TOG)}, vol.~38, no.~4, pp. 1--15, 2019.

\bibitem{hasson2019learning}
Y.~Hasson, G.~Varol, D.~Tzionas, I.~Kalevatykh, M.~J. Black, I.~Laptev, and
  C.~Schmid, ``Learning joint reconstruction of hands and manipulated
  objects,'' in \emph{Proceedings of the IEEE Conference on Computer Vision and
  Pattern Recognition}, 2019, pp. 11\,807--11\,816.

\bibitem{bernardin2005sensor}
K.~Bernardin, K.~Ogawara, K.~Ikeuchi, and R.~Dillmann, ``A sensor fusion
  approach for recognizing continuous human grasping sequences using hidden
  markov models,'' \emph{IEEE Transactions on Robotics}, vol.~21, no.~1, pp.
  47--57, 2005.

\bibitem{brahmbhatt2019contactdb}
S.~Brahmbhatt, C.~Ham, C.~C. Kemp, and J.~Hays, ``Contactdb: Analyzing and
  predicting grasp contact via thermal imaging,'' in \emph{Proceedings of the
  IEEE Conference on Computer Vision and Pattern Recognition}, 2019, pp.
  8709--8719.

\bibitem{brahmbhatt2020contactpose}
S.~Brahmbhatt, C.~Tang, C.~D. Twigg, C.~C. Kemp, and J.~Hays, ``Contactpose: A
  dataset of grasps with object contact and hand pose,'' \emph{arXiv preprint
  arXiv:2007.09545}, 2020.

\bibitem{gandarias2019cnn}
J.~M. Gandarias, A.~J. Garcia-Cerezo, and J.~M. Gomez-de Gabriel, ``Cnn-based
  methods for object recognition with high-resolution tactile sensors,''
  \emph{IEEE Sensors Journal}, vol.~19, no.~16, pp. 6872--6882, 2019.

\bibitem{corradi2015bayesian}
T.~Corradi, P.~Hall, and P.~Iravani, ``Bayesian tactile object recognition:
  Learning and recognising objects using a new inexpensive tactile sensor,'' in
  \emph{2015 IEEE international conference on robotics and automation
  (ICRA)}.\hskip 1em plus 0.5em minus 0.4em\relax IEEE, 2015, pp. 3909--3914.

\bibitem{bauza2019tactile}
M.~Bauza, O.~Canal, and A.~Rodriguez, ``Tactile mapping and localization from
  high-resolution tactile imprints,'' in \emph{2019 International Conference on
  Robotics and Automation (ICRA)}.\hskip 1em plus 0.5em minus 0.4em\relax IEEE,
  2019, pp. 3811--3817.

\bibitem{erickson2017semi}
Z.~Erickson, S.~Chernova, and C.~C. Kemp, ``Semi-supervised haptic material
  recognition for robots using generative adversarial networks,'' \emph{arXiv
  preprint arXiv:1707.02796}, 2017.

\bibitem{zheng2019bio}
W.~Zheng, B.~Wang, H.~Liu, X.~Wang, Y.~Li, and C.~Zhang, ``Bio-inspired
  magnetostrictive tactile sensor for surface material recognition,''
  \emph{IEEE Transactions on Magnetics}, vol.~55, no.~7, pp. 1--7, 2019.

\bibitem{wang2020recognition}
Y.~Wang, J.~Chen, and D.~Mei, ``Recognition of surface texture with wearable
  tactile sensor array: A pilot study,'' \emph{Sensors and Actuators A:
  Physical}, p. 111972, 2020.

\bibitem{Yuan_2017_CVPR}
W.~Yuan, S.~Wang, S.~Dong, and E.~Adelson, ``Connecting look and feel:
  Associating the visual and tactile properties of physical materials,'' in
  \emph{Proceedings of the IEEE Conference on Computer Vision and Pattern
  Recognition (CVPR)}, July 2017.

\bibitem{Li_2019_CVPR}
Y.~Li, J.-Y. Zhu, R.~Tedrake, and A.~Torralba, ``Connecting touch and vision
  via cross-modal prediction,'' in \emph{The IEEE Conference on Computer Vision
  and Pattern Recognition (CVPR)}, June 2019.

\bibitem{lee2019touching}
J.-T. Lee, D.~Bollegala, and S.~Luo, ``“touching to see” and “seeing to
  feel”: Robotic cross-modal sensory data generation for visual-tactile
  perception,'' in \emph{2019 International Conference on Robotics and
  Automation (ICRA)}.\hskip 1em plus 0.5em minus 0.4em\relax IEEE, 2019, pp.
  4276--4282.

\bibitem{tian2019manipulation}
S.~Tian, F.~Ebert, D.~Jayaraman, M.~Mudigonda, C.~Finn, R.~Calandra, and
  S.~Levine, ``Manipulation by feel: Touch-based control with deep predictive
  models,'' in \emph{2019 International Conference on Robotics and Automation
  (ICRA)}.\hskip 1em plus 0.5em minus 0.4em\relax IEEE, 2019, pp. 818--824.

\bibitem{she2019cable}
Y.~She, S.~Wang, S.~Dong, N.~Sunil, A.~Rodriguez, and E.~Adelson, ``Cable
  manipulation with a tactile-reactive gripper,'' \emph{arXiv preprint
  arXiv:1910.02860}, 2019.

\bibitem{narang2020interpreting}
Y.~S. Narang, K.~Van~Wyk, A.~Mousavian, and D.~Fox, ``Interpreting and
  predicting tactile signals via a physics-based and data-driven framework,''
  \emph{arXiv preprint arXiv:2006.03777}, 2020.

\bibitem{kim2014catching}
S.~Kim, A.~Shukla, and A.~Billard, ``Catching objects in flight,'' \emph{IEEE
  Transactions on Robotics}, vol.~30, no.~5, pp. 1049--1065, 2014.

\bibitem{akkaya2019solving}
I.~Akkaya, M.~Andrychowicz, M.~Chociej, M.~Litwin, B.~McGrew, A.~Petron,
  A.~Paino, M.~Plappert, G.~Powell, R.~Ribas, \emph{et~al.}, ``Solving rubik's
  cube with a robot hand,'' \emph{arXiv preprint arXiv:1910.07113}, 2019.

\bibitem{andrychowicz2020learning}
O.~M. Andrychowicz, B.~Baker, M.~Chociej, R.~Jozefowicz, B.~McGrew,
  J.~Pachocki, A.~Petron, M.~Plappert, G.~Powell, A.~Ray, \emph{et~al.},
  ``Learning dexterous in-hand manipulation,'' \emph{The International Journal
  of Robotics Research}, vol.~39, no.~1, pp. 3--20, 2020.

\bibitem{nagabandi2020deep}
A.~Nagabandi, K.~Konolige, S.~Levine, and V.~Kumar, ``Deep dynamics models for
  learning dexterous manipulation,'' in \emph{Conference on Robot
  Learning}.\hskip 1em plus 0.5em minus 0.4em\relax PMLR, 2020, pp. 1101--1112.

\bibitem{cutkosky2008force}
M.~R. Cutkosky, R.~D. Howe, and W.~R. Provancher, ``Force and tactile
  sensors.'' \emph{Springer Handbook of Robotics}, vol. 100, pp. 455--476,
  2008.

\bibitem{lederman2009haptic}
S.~J. Lederman and R.~L. Klatzky, ``Haptic perception: A tutorial,''
  \emph{Attention, Perception, \& Psychophysics}, vol.~71, no.~7, pp.
  1439--1459, 2009.

\bibitem{johnson2009retrographic}
M.~K. Johnson and E.~H. Adelson, ``Retrographic sensing for the measurement of
  surface texture and shape,'' in \emph{2009 IEEE Conference on Computer Vision
  and Pattern Recognition}.\hskip 1em plus 0.5em minus 0.4em\relax IEEE, 2009,
  pp. 1070--1077.

\bibitem{johnson2011microgeometry}
M.~K. Johnson, F.~Cole, A.~Raj, and E.~H. Adelson, ``Microgeometry capture
  using an elastomeric sensor,'' \emph{ACM Transactions on Graphics (TOG)},
  vol.~30, no.~4, pp. 1--8, 2011.

\bibitem{kappassov2015tactile}
Z.~Kappassov, J.-A. Corrales, and V.~Perdereau, ``Tactile sensing in dexterous
  robot hands,'' \emph{Robotics and Autonomous Systems}, vol.~74, pp. 195--220,
  2015.

\bibitem{yuan2017gelsight}
W.~Yuan, S.~Dong, and E.~H. Adelson, ``Gelsight: High-resolution robot tactile
  sensors for estimating geometry and force,'' \emph{Sensors}, vol.~17, no.~12,
  p. 2762, 2017.

\bibitem{yuan2016estimating}
W.~Yuan, M.~A. Srinivasan, and E.~H. Adelson, ``Estimating object hardness with
  a gelsight touch sensor,'' in \emph{2016 IEEE/RSJ International Conference on
  Intelligent Robots and Systems (IROS)}.\hskip 1em plus 0.5em minus
  0.4em\relax IEEE, 2016, pp. 208--215.

\bibitem{yuan2017shape}
W.~Yuan, C.~Zhu, A.~Owens, M.~A. Srinivasan, and E.~H. Adelson,
  ``Shape-independent hardness estimation using deep learning and a gelsight
  tactile sensor,'' in \emph{2017 IEEE International Conference on Robotics and
  Automation (ICRA)}.\hskip 1em plus 0.5em minus 0.4em\relax IEEE, 2017, pp.
  951--958.

\bibitem{liu2017glove}
H.~Liu, X.~Xie, M.~Millar, M.~Edmonds, F.~Gao, Y.~Zhu, V.~J. Santos,
  B.~Rothrock, and S.-C. Zhu, ``A glove-based system for studying hand-object
  manipulation via joint pose and force sensing,'' in \emph{2017 IEEE/RSJ
  International Conference on Intelligent Robots and Systems (IROS)}.\hskip 1em
  plus 0.5em minus 0.4em\relax IEEE, 2017, pp. 6617--6624.

\bibitem{liu2019high}
H.~Liu, Z.~Zhang, X.~Xie, Y.~Zhu, Y.~Liu, Y.~Wang, and S.-C. Zhu,
  ``High-fidelity grasping in virtual reality using a glove-based system,'' in
  \emph{2019 International Conference on Robotics and Automation (ICRA)}.\hskip
  1em plus 0.5em minus 0.4em\relax IEEE, 2019, pp. 5180--5186.

\bibitem{luo2021learning}
Y.~Luo, Y.~Li, P.~Sharma, W.~Shou, K.~Wu, M.~Foshey, B.~Li, T.~Palacios,
  A.~Torralba, and W.~Matusik, ``Learning human--environment interactions using
  conformal tactile textiles,'' \emph{Nature Electronics}, vol.~4, no.~3, pp.
  193--201, 2021.

\bibitem{lee2019making}
M.~A. Lee, Y.~Zhu, K.~Srinivasan, P.~Shah, S.~Savarese, L.~Fei-Fei, A.~Garg,
  and J.~Bohg, ``Making sense of vision and touch: Self-supervised learning of
  multimodal representations for contact-rich tasks,'' in \emph{2019
  International Conference on Robotics and Automation (ICRA)}.\hskip 1em plus
  0.5em minus 0.4em\relax IEEE, 2019, pp. 8943--8950.

\bibitem{veiga2018grip}
F.~Veiga, J.~Peters, and T.~Hermans, ``Grip stabilization of novel objects
  using slip prediction,'' \emph{IEEE transactions on haptics}, vol.~11, no.~4,
  pp. 531--542, 2018.

\bibitem{dong2019maintaining}
S.~Dong, D.~Ma, E.~Donlon, and A.~Rodriguez, ``Maintaining grasps within
  slipping bounds by monitoring incipient slip,'' in \emph{2019 International
  Conference on Robotics and Automation (ICRA)}.\hskip 1em plus 0.5em minus
  0.4em\relax IEEE, 2019, pp. 3818--3824.

\bibitem{chebotar2016self}
Y.~Chebotar, K.~Hausman, Z.~Su, G.~S. Sukhatme, and S.~Schaal,
  ``Self-supervised regrasping using spatio-temporal tactile features and
  reinforcement learning,'' in \emph{2016 IEEE/RSJ International Conference on
  Intelligent Robots and Systems (IROS)}.\hskip 1em plus 0.5em minus
  0.4em\relax IEEE, 2016, pp. 1960--1966.

\bibitem{hogan2018tactile}
F.~R. Hogan, M.~Bauza, O.~Canal, E.~Donlon, and A.~Rodriguez, ``Tactile
  regrasp: Grasp adjustments via simulated tactile transformations,'' in
  \emph{2018 IEEE/RSJ International Conference on Intelligent Robots and
  Systems (IROS)}.\hskip 1em plus 0.5em minus 0.4em\relax IEEE, 2018, pp.
  2963--2970.

\bibitem{frome2013devise}
A.~Frome, G.~S. Corrado, J.~Shlens, S.~Bengio, J.~Dean, M.~Ranzato, and
  T.~Mikolov, ``Devise: A deep visual-semantic embedding model,'' in
  \emph{Advances in neural information processing systems}, 2013, pp.
  2121--2129.

\bibitem{karpathy2015deep}
A.~Karpathy and L.~Fei-Fei, ``Deep visual-semantic alignments for generating
  image descriptions,'' in \emph{Proceedings of the IEEE conference on computer
  vision and pattern recognition}, 2015, pp. 3128--3137.

\bibitem{xu2015show}
K.~Xu, J.~Ba, R.~Kiros, K.~Cho, A.~Courville, R.~Salakhudinov, R.~Zemel, and
  Y.~Bengio, ``Show, attend and tell: Neural image caption generation with
  visual attention,'' in \emph{International conference on machine learning},
  2015, pp. 2048--2057.

\bibitem{norouzi2013zero}
M.~Norouzi, T.~Mikolov, S.~Bengio, Y.~Singer, J.~Shlens, A.~Frome, G.~S.
  Corrado, and J.~Dean, ``Zero-shot learning by convex combination of semantic
  embeddings,'' \emph{arXiv preprint arXiv:1312.5650}, 2013.

\bibitem{otani2016learning}
M.~Otani, Y.~Nakashima, E.~Rahtu, J.~Heikkil{\"a}, and N.~Yokoya, ``Learning
  joint representations of videos and sentences with web image search,'' in
  \emph{European Conference on Computer Vision}.\hskip 1em plus 0.5em minus
  0.4em\relax Springer, 2016, pp. 651--667.

\bibitem{aytar2017cross}
Y.~Aytar, L.~Castrejon, C.~Vondrick, H.~Pirsiavash, and A.~Torralba,
  ``Cross-modal scene networks,'' \emph{IEEE transactions on pattern analysis
  and machine intelligence}, vol.~40, no.~10, pp. 2303--2314, 2017.

\bibitem{owens2016ambient}
A.~Owens, J.~Wu, J.~H. McDermott, W.~T. Freeman, and A.~Torralba, ``Ambient
  sound provides supervision for visual learning,'' in \emph{European
  conference on computer vision}.\hskip 1em plus 0.5em minus 0.4em\relax
  Springer, 2016, pp. 801--816.

\bibitem{gan2020foley}
C.~Gan, D.~Huang, P.~Chen, J.~B. Tenenbaum, and A.~Torralba, ``Foley music:
  Learning to generate music from videos,'' \emph{arXiv preprint
  arXiv:2007.10984}, 2020.

\bibitem{zhao2019sound}
H.~Zhao, C.~Gan, W.-C. Ma, and A.~Torralba, ``The sound of motions,'' in
  \emph{Proceedings of the IEEE International Conference on Computer Vision},
  2019, pp. 1735--1744.

\bibitem{falco2017cross}
P.~Falco, S.~Lu, A.~Cirillo, C.~Natale, S.~Pirozzi, and D.~Lee, ``Cross-modal
  visuo-tactile object recognition using robotic active exploration,'' in
  \emph{2017 IEEE International Conference on Robotics and Automation
  (ICRA)}.\hskip 1em plus 0.5em minus 0.4em\relax IEEE, 2017, pp. 5273--5280.

\bibitem{luo2021intelligent}
Y.~Luo, Y.~Li, M.~Foshey, W.~Shou, P.~Sharma, T.~Palacios, A.~Torralba, and
  W.~Matusik, ``Intelligent carpet: Inferring 3d human pose from tactile
  signals,'' in \emph{Proceedings of the IEEE/CVF Conference on Computer Vision
  and Pattern Recognition}, 2021, pp. 11\,255--11\,265.

\bibitem{luo2019iclap}
S.~Luo, W.~Mou, K.~Althoefer, and H.~Liu, ``iclap: Shape recognition by
  combining proprioception and touch sensing,'' \emph{Autonomous Robots},
  vol.~43, no.~4, pp. 993--1004, 2019.

\bibitem{pastor2020bayesian}
F.~Pastor, J.~Garc{\'\i}a-Gonz{\'a}lez, J.~M. Gandarias, D.~Medina, P.~Closas,
  A.~J. Garc{\'\i}a-Cerezo, and J.~M. G{\'o}mez-de Gabriel, ``Bayesian and
  neural inference on lstm-based object recognition from tactile and
  kinesthetic information,'' \emph{IEEE Robotics and Automation Letters},
  vol.~6, no.~1, pp. 231--238, 2020.

\bibitem{oikonomidis2011full}
I.~Oikonomidis, N.~Kyriazis, and A.~A. Argyros, ``Full dof tracking of a hand
  interacting with an object by modeling occlusions and physical constraints,''
  in \emph{2011 International Conference on Computer Vision}.\hskip 1em plus
  0.5em minus 0.4em\relax IEEE, 2011, pp. 2088--2095.

\bibitem{sridhar2016real}
S.~Sridhar, F.~Mueller, M.~Zollh{\"o}fer, D.~Casas, A.~Oulasvirta, and
  C.~Theobalt, ``Real-time joint tracking of a hand manipulating an object from
  rgb-d input,'' in \emph{European Conference on Computer Vision}.\hskip 1em
  plus 0.5em minus 0.4em\relax Springer, 2016, pp. 294--310.

\bibitem{kyriazis2014scalable}
N.~Kyriazis and A.~Argyros, ``Scalable 3d tracking of multiple interacting
  objects,'' in \emph{Proceedings of the IEEE Conference on Computer Vision and
  Pattern Recognition}, 2014, pp. 3430--3437.

\bibitem{pham2015towards}
T.-H. Pham, A.~Kheddar, A.~Qammaz, and A.~A. Argyros, ``Towards force sensing
  from vision: Observing hand-object interactions to infer manipulation
  forces,'' in \emph{Proceedings of the IEEE conference on computer vision and
  pattern recognition}, 2015, pp. 2810--2819.

\bibitem{tzionas2016capturing}
D.~Tzionas, L.~Ballan, A.~Srikantha, P.~Aponte, M.~Pollefeys, and J.~Gall,
  ``Capturing hands in action using discriminative salient points and physics
  simulation,'' \emph{International Journal of Computer Vision}, vol. 118,
  no.~2, pp. 172--193, 2016.

\bibitem{pham2017hand}
T.-H. Pham, N.~Kyriazis, A.~A. Argyros, and A.~Kheddar, ``Hand-object contact
  force estimation from markerless visual tracking,'' \emph{IEEE transactions
  on pattern analysis and machine intelligence}, vol.~40, no.~12, pp.
  2883--2896, 2017.

\bibitem{ehsani2020use}
K.~Ehsani, S.~Tulsiani, S.~Gupta, A.~Farhadi, and A.~Gupta, ``Use the force,
  luke! learning to predict physical forces by simulating effects,'' in
  \emph{Proceedings of the IEEE/CVF Conference on Computer Vision and Pattern
  Recognition}, 2020, pp. 224--233.

\bibitem{chopra2005learning}
S.~Chopra, R.~Hadsell, and Y.~LeCun, ``Learning a similarity metric
  discriminatively, with application to face verification,'' in \emph{2005 IEEE
  Computer Society Conference on Computer Vision and Pattern Recognition
  (CVPR'05)}, vol.~1.\hskip 1em plus 0.5em minus 0.4em\relax IEEE, 2005, pp.
  539--546.

\bibitem{schroff2015facenet}
F.~Schroff, D.~Kalenichenko, and J.~Philbin, ``Facenet: A unified embedding for
  face recognition and clustering,'' in \emph{Proceedings of the IEEE
  conference on computer vision and pattern recognition}, 2015, pp. 815--823.

\bibitem{hadsell2006dimensionality}
R.~Hadsell, S.~Chopra, and Y.~LeCun, ``Dimensionality reduction by learning an
  invariant mapping,'' in \emph{2006 IEEE Computer Society Conference on
  Computer Vision and Pattern Recognition (CVPR'06)}, vol.~2.\hskip 1em plus
  0.5em minus 0.4em\relax IEEE, 2006, pp. 1735--1742.

\bibitem{gutmann2010noise}
M.~Gutmann and A.~Hyv{\"a}rinen, ``Noise-contrastive estimation: A new
  estimation principle for unnormalized statistical models,'' in
  \emph{Proceedings of the Thirteenth International Conference on Artificial
  Intelligence and Statistics}, 2010, pp. 297--304.

\bibitem{wu2018unsupervised}
Z.~Wu, Y.~Xiong, S.~X. Yu, and D.~Lin, ``Unsupervised feature learning via
  non-parametric instance discrimination,'' in \emph{Proceedings of the IEEE
  Conference on Computer Vision and Pattern Recognition}, 2018, pp. 3733--3742.

\bibitem{oord2018representation}
A.~v.~d. Oord, Y.~Li, and O.~Vinyals, ``Representation learning with
  contrastive predictive coding,'' \emph{arXiv preprint arXiv:1807.03748},
  2018.

\bibitem{sun2014deep}
Y.~Sun, Y.~Chen, X.~Wang, and X.~Tang, ``Deep learning face representation by
  joint identification-verification,'' in \emph{Advances in neural information
  processing systems}, 2014, pp. 1988--1996.

\bibitem{zhang2016embedding}
X.~Zhang, F.~Zhou, Y.~Lin, and S.~Zhang, ``Embedding label structures for
  fine-grained feature representation,'' in \emph{Proceedings of the IEEE
  Conference on Computer Vision and Pattern Recognition}, 2016, pp. 1114--1123.

\bibitem{koenker2001quantile}
R.~Koenker and K.~F. Hallock, ``Quantile regression,'' \emph{Journal of
  economic perspectives}, vol.~15, no.~4, pp. 143--156, 2001.

\bibitem{huttenlocher1993comparing}
D.~P. Huttenlocher, G.~A. Klanderman, and W.~J. Rucklidge, ``Comparing images
  using the hausdorff distance,'' \emph{IEEE Transactions on pattern analysis
  and machine intelligence}, vol.~15, no.~9, pp. 850--863, 1993.

\bibitem{Lukezic_IJCV2018}
A.~Luke{\v{z}}i{\v{c}}, T.~Voj{'i}{\v{r}}, L.~{\v{C}}ehovin~Zajc, J.~Matas, and
  M.~Kristan, ``Discriminative correlation filter tracker with channel and
  spatial reliability,'' \emph{International Journal of Computer Vision}, 2018.

\bibitem{kingma2014adam}
D.~P. Kingma and J.~Ba, ``Adam: A method for stochastic optimization,''
  \emph{arXiv preprint arXiv:1412.6980}, 2014.

\end{thebibliography}
